\documentclass[11pt,a4paper]{article}
\usepackage{times,latexsym}
\usepackage{url}
\usepackage[T1]{fontenc}

\usepackage[acceptedWithA]{tacl2021v1}

\usepackage{xspace,mfirstuc,tabulary}

\newif\iftaclinstructions
\taclinstructionsfalse 
\iftaclinstructions
\renewcommand{\confidential}{}
\renewcommand{\anonsubtext}{(No author info supplied here, for consistency with
TACL-submission anonymization requirements)}
\newcommand{\instr}
\fi

\iftaclpubformat 

\else

\fi

\usepackage{suffix}
\usepackage{amsmath}
\usepackage{booktabs}
\usepackage{multirow}
\usepackage{graphicx}
\usepackage{xcolor}
\usepackage[para]{threeparttable}
\usepackage{tabularx}
\newcolumntype{Y}{>{\centering\arraybackslash}X}
\newcolumntype{Z}{>{\raggedleft\arraybackslash}X}

\newcommand{\note}[2]{\textmd{\textit{\textcolor{#1}{*#2}}}}
\WithSuffix\newcommand\note*[2]{\textmd{\textit{\textcolor{#1}{*#2}}\hfill\break}}
\newcommand{\warning}[1]{\note{orange}{#1}}
\WithSuffix\newcommand\warning*[1]{\note*{orange}{#1}}
\newcommand{\error}[1]{\note{red}{#1}}
\WithSuffix\newcommand\error*[1]{\note*{red}{#1}}

\newcommand{\fillin}{\error{Fill in.}}
\WithSuffix\newcommand\fillin*{\error*{Fill in.}}
\newif\ifguide
\guidetrue

\newcommand{\refhere}[1]{\warning{reference here}\xspace\ }

\definecolor{mygreen}{rgb}{0.0, 0.5, 0.0}
\newcommand{\revb}[1]{\textmd{#1}}
\newcommand{\rev}[1]{\textmd{#1}}

\newcommand{\pair}[1]{En$\to$#1}
\newcommand{\pattern}[3]{\texttt{#1\textasciitilde #2\textasciitilde #3}}
\newcommand{\defA}{word-based}
\newcommand{\defB}{arc-based}
\newcommand{\defACap}{Word-based}
\newcommand{\defBCap}{Arc-based}

\newcommand{\mc}{\multicolumn}
\newcommand{\mr}{\multirow}
\newcommand{\biasDescSingular}{\rev{appears around 50\% of the time in training data}}
\newcommand{\biasDescPlural}{\rev{appear around 50\% of the time in training data}}
\newcommand{\htprob}{$\Pr_{\text{HT}}(\cdot\mid p)$\xspace}
\newcommand{\mtprob}{$\Pr_{\text{MT}}(\cdot\mid p)$\xspace}

\title{To Diverge or Not to Diverge: A Morphosyntactic Perspective on Machine Translation vs Human Translation}

\author{
  Jiaming Luo
  \and
  Colin Cherry
  \and
  George Foster
  \\
  \ \\
  Google Translate Research
  \\
  \texttt{\{jmluo,colincherry,fosterg\}.google.com}
}

\date{}

\begin{document}
\maketitle

\begin{abstract}
    We conduct a large-scale fine-grained comparative analysis
    of machine translations (MT) against human translations (HT) through
    the lens of morphosyntactic divergence.
    Across three language pairs and two \revb{types} of divergence
    \revb{defined as the structural difference between the source and the target,}
    MT is consistently more conservative than HT, with less morphosyntactic diversity, more convergent patterns, and more one-to-one alignments.
    Through analysis on different decoding algorithms,
    we attribute this discrepancy to the use of beam search
    that biases MT towards more convergent patterns.
    This bias is most amplified when the convergent pattern \biasDescSingular{}.
    Lastly, we show that for a majority of morphosyntactic divergences,
    their presence in HT is
    correlated with decreased MT performance, presenting a greater challenge for MT systems.
\end{abstract}
\section{Introduction}
\label{sec:intro}

\newcommand{\ul}[1]{\underline{#1}}
\newcommand{\rootDep}[1]{\textbf{#1}}
\newcommand{\xcompDep}[1]{\textbf{#1}}

\begin{figure}[ht!]
    \centering
    \scriptsize
    \begin{tabular}{l}
        \toprule

        Readers are \rootDep{cautioned} not to \xcompDep{place} undue reliance on \ldots \\
        \ul{Il} est recommand\'e aux \ul{lecteurs} de ne pas accorder\ldots \\
        \midrule
                
        PCO has \rootDep{continued} to \xcompDep{assist} \ldots \\
        Le BCP a \ul{constamment} \'epaul\'e \ldots \\
        \midrule
        

        
        
        
        
        
        These meetings \rootDep{seek} to \xcompDep{strengthen} \ldots \\
        Ces r\'eunions \ul{ont pour but} de renforcer \ldots \\
        \midrule
        
        \ldots weaknesses will \rootDep{become} more \xcompDep{apparent} \ldots \\
        Les faiblesses \ul{appara\^itront}\ldots \\

        \bottomrule
    \end{tabular}

    \vspace{0.3cm}
    \includegraphics[width=\linewidth]{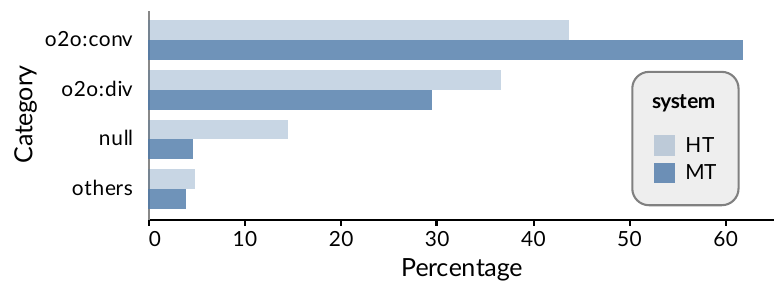}
    \caption{\textbf{Top table}: Examples of divergences
        in HT for \pair{Fr} \revb{WMT15 training data~\citep{bojar-etal-2015-findings}}, with relevant fragments of the source/target shown in the first/second rows.
        The English control constructions are bolded including both the finite root verb and the controlled word, 
        while the French phrases of interest are underlined.
        \textbf{Bottom figure}: Percentages of target patterns for HT and MT, with obligatory control finite verbs as the source pattern. 
        \texttt{o2o:conv}: one-to-one convergent patterns \revb{where the target phrase uses a similar control construction to the source};
        \texttt{o2o:div}: one-to-one divergent patterns \revb{where the target differs structurally from the source};
        \texttt{null}: no target word is aligned;
        \texttt{others}: other less frequent patterns \revb{(e.g., one-to-many alignments)}. 
        \revb{The percentages of all four categories sum up to 100\%.}
        }
    \label{fig:example}
\end{figure}

Translation divergences occur when the translations differ structurally from the source sentences,
typically as a result of either inherent crosslingual differences
or idiosyncratic preferences of translators.
\revb{These divergences happen naturally in the translation process and can be readily found in human translations (HT), including those used for training machine translation (MT) systems (see the table in Figure~\ref{fig:example} for some examples).}
Their existence in \revb{HT} has long been regarded
as a key challenge for \revb{MT}~\cite{Dorr1994MachineTD}
and more recent empirical studies have demonstrated the abundance of translation divergences in
HT ~\cite{Deng2017TranslationDI,Nikolaev2020FineGrainedAO}.

\revb{
In contrast to HT, MT outputs tend to be less diverse and more literal (i.e., absence of translation divergence), exhibiting the features of \emph{translationese}~\citep{gellerstam1986translationese}.
This \emph{qualitative} difference between HT and MT has inspired a rich body of work attempting to narrow the gap, such as automatic
detection of machine translated texts in the training data~\citep{kurokawa-etal-2009-automatic,lembersky-etal-2012-adapting,aharoni-etal-2014-automatic,riley-etal-2020-translationese,freitag-etal-2022-natural},
training MT systems on more diverse translations~\citep{khayrallah-etal-2020-simulated,bao-etal-2023-target}, and carefully reordering the examples to reduce the degree of divergence between the source and the target~\citep{wang-etal-2007-chinese,zhang-zong-2016-exploiting,zhou-etal-2019-handling}.
The challenges that translation divergences present do not just concern training MT systems, but also their evaluation~\citep{koppel-ordan-2011-translationese,freitag-etal-2020-bleu}.}

\revb{
Nonetheless, even } as we gain deepened understanding of \revb{how to address these challenges}, it remains unclear 
 \emph{how quantitatively different} MT and HT are in terms of divergences.\footnote{We use the term MT to mean the version of MT tested in this project's experiments: bilingual encoder-decoder Transformer-base networks with beam search decoding (see Models in Section~\ref{sec:experimental_setup}).}
Control verbs,\footnote{\url{https://en.wikipedia.org/wiki/Control_(linguistics)}. They are coded as \texttt{xcomp} in Universal Dependencies (see \url{https://universaldependencies.org/u/dep/all.html\#xcomp-open-clausal-complement}).}  for instance, provide a great case study
to showcase this difference.
There is much uncertainty when translating them from English to French, and human translators employ a wide variety of constructions including many divergent patterns (Figure~\ref{fig:example}).
In comparison,
MT is much more likely to preserve the source structure, with the convergent pattern comprising about 20\% more of all translations of control verbs.
This difference exemplifies MT's undesirable tendency to
produce translationese that is too literal and lacks structural diversity~\cite{freitag-etal-2019-ape,bizzoni-etal-2020-human}.

In this work, we seek to systematically investigate this difference
by conducting a \emph{large-scale} \emph{fine-grained} \emph{comparative} analysis
on the distribution of translation divergences for HT and MT,
all through the lens of morphosyntax.
More specifically, we aim to answer the following research questions:
1) How are MT and HT quantitatively different in terms of morphosyntactic divergence?
2) How do we explain or understand this difference?
3) How do translation divergences in HT affect MT quality? In other words, do MT systems have
more difficulty translating source sentences that exhibit divergences in HT?


Through extensive analyses based on three language pairs and two \revb{types} of morphosyntactic divergence
using the annotational framework of Universal Dependencies~\cite{nivre2016universal},
we make the following empirical observations:
\begin{enumerate}
    \item MT is more \emph{conservative} than HT, with less morphosyntactic diversity, more convergent patterns, and more one-to-one alignments.
    \item MT is morphosyntactically less similar to HT for less frequent source patterns.
    \item The distributional difference can be largely attributed to the use of beam search, which
          is biased towards convergent patterns. This bias is most amplified when the convergent
         \revb{target patterns appear around 50\% of the time out of all translations of the same source pattern in the training data}.
    \item A majority of the most frequent divergent patterns are correlated with decreased MT performance. This correlation cannot be fully explained by the lower frequencies
          of the relevant divergences.
\end{enumerate}

\revb{To the best of our knowledge, this is the first work to present the comparative perspective of HT vs MT in such fine granularity covering thousands of morphosyntactic constructions.} In the remaining sections, we first briefly describe related work in Section~\ref{sec:related_work}.
The experimental setup is described in detail in Section~\ref{sec:experimental_setup}.
We demonstrate the quantitative difference between MT and HT in Section~\ref{sec:comparison},
and seek to understand this discrepancy in Section~\ref{sec:understanding}.
Lastly, we explore the correlation between the presence of divergences in HT with MT performance
in Section~\ref{sec:difficulty} and make conclusions in Section~\ref{sec:conclusion}.
\section{Related Work}
\label{sec:related_work}

\paragraph{Translation Divergence}
Systematic and theoretical treatment of translation divergences started in the
early 1990s, focusing on European languages~\cite{dorr1992use,dorr1993interlingual,Dorr1994MachineTD}.
Later work has expanded into more languages,
and focused on the automatic detection of divergences~\cite{gupta2001study,Gupta2003IdentificationOD,Sinha2005TranslationDI,Mishra2009DivergencePB,Saboor2010LexicalsemanticDI}
or their empirical distributions in human
translations~\cite{wong2017quantitative,Deng2017TranslationDI,Wein2021ClassifyingDI}.
\revb{Relatedly, \citet{carpuat-etal-2017-detecting,vyas-etal-2018-identifying,briakou-carpuat-2020-detecting} focused on identifying semantic divergences that manifest in translations not entirely semantically equivalent to the original sources.}

The closest work to ours is from \citet{Nikolaev2020FineGrainedAO} who
proposed to investigate fine-grained crosslingual morphosyntactic divergence
based on Universal Dependencies.
They augmented a subset of the Parallel Universal Dependencies (PUD) corpus~\cite{zeman-etal-2017-conll}
with human-annotated word alignments for five language pairs and focused
exclusively on content words.
While our work shares a similar conceptional and methodological
foundation to theirs, our goal is to conduct a comparative analysis between HT and MT.
In addition, we rely on \revb{a dependency parser and a word aligner~(see Section~\ref{sec:experimental_setup} for more details)} to reach a sufficiently large scale
to enable the investigation of more fine-grained divergences.

\paragraph{Diverse Machine Translation}
MT systems tend to produce less diverse outputs in general~\cite{gimpel2013systematic,ott2018analyzing},
which is particularly harmful for back translation~\cite{edunov-etal-2018-understanding,soto-etal-2020-selecting,burchell-birch-and-kenneth-heafield-2022-exploring}.
To address this issue, various techniques have been proposed in the literature,
including modified decoding algorithms~\cite{Li2016ASF,sun2020generating,li2021mixup},
mixtures of experts~\cite{shen2019mixture},
Bayesian models~\cite{wu2020generating}, additional codes (syntax or latent)~\cite{shu-etal-2019-generating,lachaux-etal-2020-target}
and training with simulated multi-reference corpora~\cite{lin-etal-2022-bridging}.
In all aforementioned works, the emphasis is on the lack of diversity in MT outputs rather than comparing 
them systematically against HT.
Notable exceptions include~\citet{Roberts2020} who investigated the distributional
differences between MT and HT in terms of n-grams, sentence length, punctuation, and copy rates.
\citet{marchisio-etal-2022-systematic} compared translations from supervised MT and unsupervised MT and noted their systematic style differences based on
similarity and monotonicity in their POS sequences.
In contrast, our work goes beyond surface features and focuses
on fine-grained morphosyntactic divergences.

\paragraph{Algorithmic Bias}
Another closely related line of work studies algorithmic biases
of current NLP systems,
with particular emphasis on gender and racial biases~\cite{bolukbasi2016man,caliskan2017semantics,zhao-etal-2017-men,Garg2018WordEQ}.
Specifically for MT,
researchers have focused on lexical diversity by comparing
HT against post-editese~\cite{toral-2019-post}
or MT outputs directly~\cite{vanmassenhove-etal-2019-lost};
\citet{bizzoni-etal-2020-human} have compared HT, MT and simultaneous interpreting
in terms of translationese using POS perplexity and dependency length.
Most related to our work, \citet{vanmassenhove-etal-2021-machine} have conducted an extensive comparison
between HT and MT based on a suite of lexical and morphological diversity metrics.
While our study reaches a similar conclusion that MT is less diverse than HT,
we explore morphosyntactic patterns on a more fine-grained level,
and also reveal the bias of MT (and more specifically beam search) towards
convergent structures.

\section{Experimental Setup}
\label{sec:experimental_setup}

\begin{figure}
    \includegraphics[width=\linewidth]{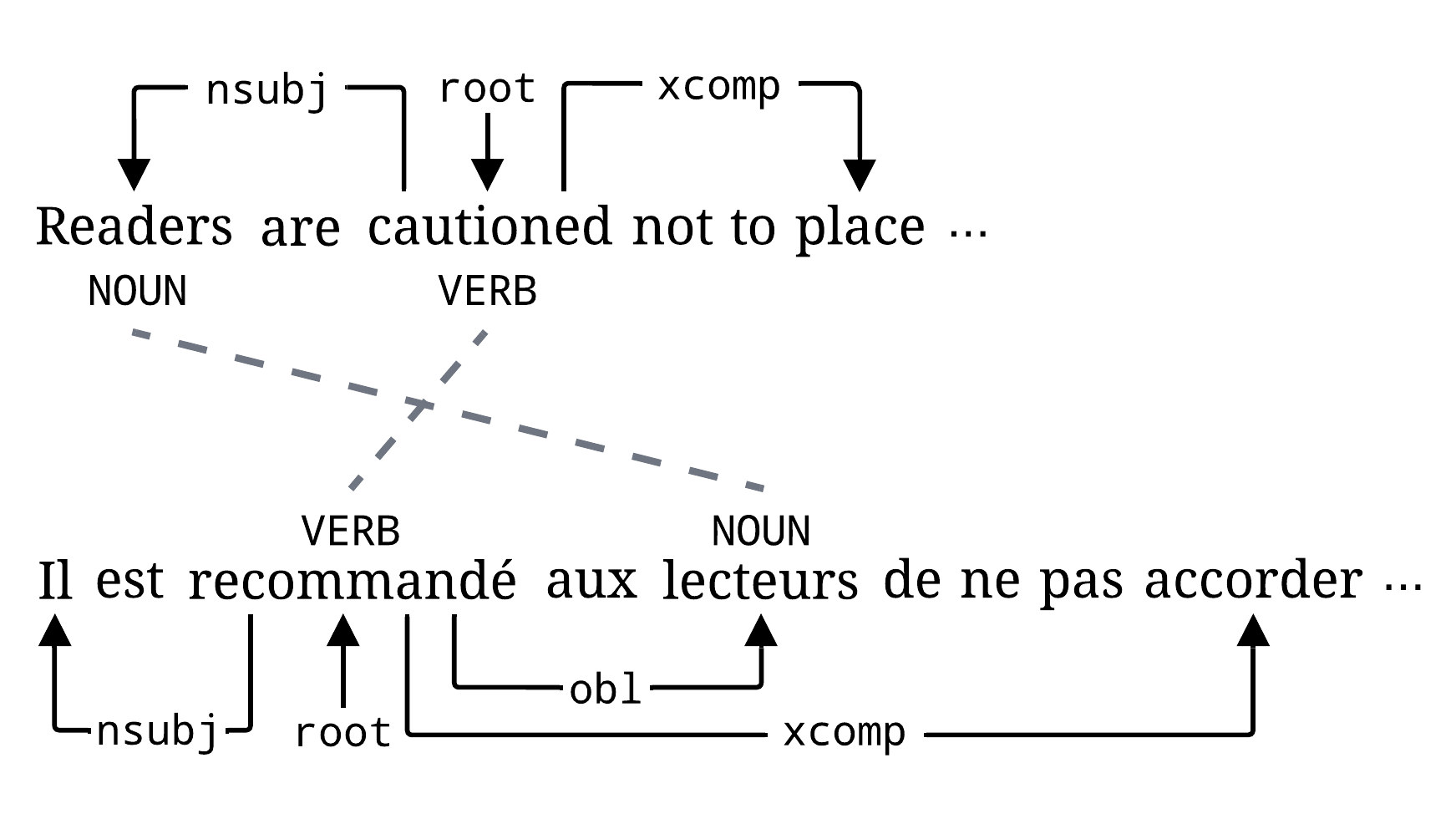}
    \caption{An illustration of the two \revb{types} of morphosyntactic
        divergence. See Section~\ref{sec:experimental_setup} for details.}
    \label{fig:divergence_example}
\end{figure}

\paragraph{\revb{Types} of Morphosyntactic Divergence}
In this study, we experiment with
two \revb{types} of translation patterns
based on the annotational scheme of Universal Dependencies:
\begin{enumerate}
    \item[(A)] \textit{\defACap{}}: POS tags for the aligned word pair.
        We additionally include their parent and child syntactic dependencies for more granularity.
        Order of the children dependencies is ignored.
    \item[(B)] \textit{\defBCap{}}: The source dependency arc, and the
        target path between the aligned words of the arc's head and tail.
        Directionality of the target dependencies is ignored.
        We additionally include the POS tags of both the head and the tail for more granularity.
\end{enumerate}
These \revb{types} are largely based on the proposal of~\citet{Nikolaev2020FineGrainedAO},
with modifications to accommodate more granularity.
With either \revb{type}, the translation pattern is a \emph{convergence}
if the source and the target sides \revb{have the same structure (\defA{} or \defB{})}, and otherwise a \emph{divergence}.
Notationally, we use tildes to connect the various parts of the pattern in a fixed order. 
For instance, for the control verb \revb{\emph{``cautioned''}}
in Figure~\ref{fig:divergence_example},
its \defA{} divergence has \pattern{root}{VERB}{nsubj+xcomp} on the source side, \revb{where \texttt{VERB} corresponds
to its POS tag, \texttt{root} its parent dependency, and 
\texttt{nsubj} and \texttt{xcomp} its two child dependencies. Similarly, we have}
\pattern{root}{VERB}{nsubj+obl+xcomp} on the target side.
With regard to an \defB{} divergence, \revb{for the source arc between the words \emph{``cautioned''} and \emph{``readers''}, we denote it as} \pattern{VERB}{nsubj}{NOUN},
\revb{where \texttt{nsubj} is the dependency relation of the arc, and \texttt{VERB} and \texttt{NOUN} the 
POS tags of the head and the tail, respectively. Similarly, we denote the aligned target pattern as}
\pattern{VERB}{obl}{NOUN}.

\paragraph{Data}
We conduct experiments for three language pairs using WMT datasets~\cite{bojar-etal-2015-findings,barrault-etal-2019-findings}: \pair{Zh} (WMT19),
\pair{Fr} (WMT15) and \pair{De} (WMT19).
All training datasets are lightly filtered based on length, length ratio and language ID,
and deduplicated.
For each language pair, one million sentences
are held out from the training split to form an analysis subset. All analyses in our study are based on this subset to eliminate potential confounding effects from domain mismatch. 
\revb{Table~\ref{tab:pattern_counts} shows the number of distinct 
source or target patterns found in the analysis set for each language pair.
\begin{table}[h]
    \centering
    \begin{tabular}{lrr}
    \toprule
    Language pair & Source & Target \\
    \midrule
    \pair{De} &17\ 055&	15\ 040 \\
    \pair{Zh} &14\ 816&	19\ 471 \\
    \pair{Fr} &18\ 321&	12\ 212 \\
    \bottomrule
    \end{tabular}
    \caption{Number of distinct source or target patterns found in the analysis set (1M sentences from WMT).}
    \label{tab:pattern_counts}
\end{table}
}

\paragraph{Models}
We train a bilingual Transformer base model~\cite{vaswani2017attention}
for each language pair using the \texttt{T5X} framework~\cite{roberts2022t5x}.
All models are trained
with \texttt{Adafactor} optimizer~\cite{shazeer2018adafactor}
for 2M steps with 0.1 dropout rate, 1024 batch size, and 0.1 label smoothing.
We use an inverse square root learning rate schedule with a base rate of 2.0.
As summarized in Table~\ref{tab:tools} part (i), all models achieve similar BLEU
scores\footnote{All reported BLEU scores for our models are obtained through \texttt{SacreBLEU}~\cite{post-2018-call}.}
on the development set as reported in the literature with a comparable setup.

\begin{table}
    \centering
    \small
    \begin{threeparttable}
        \begin{tabularx}{\linewidth}{llYY}
            \toprule
            \mc{4}{l}{(i) \textsc{Translation}}                                \\
            \midrule
            Target & Dev dataset  & BLEU & Reported                            \\
            \midrule
            Fr     & newstest2014 & 39.9 & 38.1\rlap{\tnote{$\dagger$}} \\
            De     & newstest2018 & 46.3 & 46.4\rlap{\tnote{$\ddagger$}}       \\
            Zh     & newstest2018 & 34.4 & 34.8\rlap{\tnote{$\dagger\dagger$}}        \\
        \end{tabularx}

        \begin{tabularx}{\linewidth}{llYYY}
            \midrule\midrule
            \mc{5}{l}{(ii) \textsc{Dependency Parsing}} \\
            \midrule
            Language & Dataset & UPOS  & UAS   & LAS    \\
            \midrule
            En       & EWT     & 95.56 & 89.55 & 96.67  \\
            Fr       & GSD     & 97.77 & 93.20 & 90.90  \\
            De       & GSD     & 94.80 & 87.87 & 83.25  \\
            Zh       & GSD     & 94.58 & 86.41 & 80.70  \\
        \end{tabularx}

        \begin{tabularx}{\linewidth}{lYYY}
            \midrule\midrule
            \mc{4}{l}{(iii) \textsc{Word Alignment}} \\
            \midrule
            Target & Precision & Recall & F1         \\
            \midrule
            Fr     & 85.6     & 81.7  & 83.6      \\
            Zh     & 85.5     & 81.9  & 83.7      \\
            \bottomrule
        \end{tabularx}

        \begin{tablenotes}
            \footnotesize
            \item [$\dagger$] \citet{vaswani2017attention}
            \item [$\ddagger$] \citet{ng-etal-2019-facebook}
            \item [$\dagger\dagger$] \citet{bawden-etal-2019-university}
        \end{tablenotes}
    \end{threeparttable}
    \caption{Performance for (i) MT (ii) dependency parsing and
        (iii) word alignment. No human annotations for \pair{De} are provided by \citet{Nikolaev2020FineGrainedAO}.}
    \label{tab:tools}
\end{table}

\paragraph{Annotations}
We rely on two automatic tools to conduct a large-scale 
analysis: a dependency parser and a word aligner.
More specifically, the dependency parser is an implementation of~\citet{dozat2017deep} based on mBERT~\cite{devlin-etal-2019-bert}.
The neural word aligner is based on AMBER~\cite{hu-etal-2021-explicit}
and fine-tuned on human-annotated alignments.
We follow~\citet{Nikolaev2020FineGrainedAO}
to keep the content words\footnote{\revb{Content words are words with semantic content, used in various ``contentful'' positions such as subjects, objects and adjectival modifiers.} We identify
    content words by matching their parent dependencies against a manually selected set,
    as defined in footnote 10 of the original paper~\citep{Nikolaev2020FineGrainedAO}.
 \revb{This criterion kept around 40\%-50\% of all the tokens for all three language pairs in our experiments. Please see Appendix~\ref{appendix:content_stats} for a more detailed analysis.}
} and their dependencies and alignments only,
and focus on one-to-one alignments unless otherwise noted.

As reported in Table~\ref{tab:tools} part (ii) and (iii),
we validate that both tools have high accuracy on public datasets:
UD test sets for parsing and human-annotated PUD datasets~\cite{Nikolaev2020FineGrainedAO}
for word alignment.
We will release the automatic annotations to the public.

\section{Comparative Analysis of MT vs HT}
\label{sec:comparison}

\begin{figure}[tb]
    \centering
    \includegraphics[width=\linewidth]{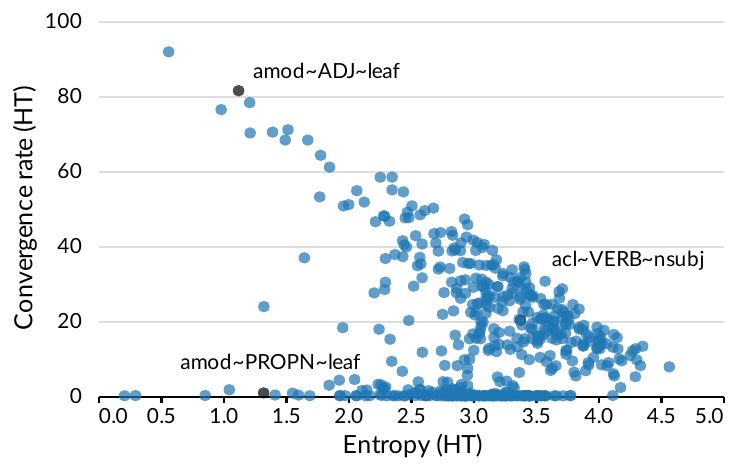}
    \caption{Plot of convergence rate vs entropy for the most frequent \defA{} source patterns in \pair{Fr} human translations, three of which are highlighted in black: 
    (1) \pattern{amod}{ADJ}{leaf} (high convergence rate, low entropy): the most common cases of adjectival modifiers; 
    (2) \pattern{acl}{VERB}{nsubj} (low convergence rate, high entropy): object relative clauses without a relative pronoun, or subject relative clauses. The high entropy reflects a major difference between English and French, where the relative pronoun \textit{que} is obligatory in French but not in English.
    (3) \pattern{amod}{PROPN}{leaf} (low convergence rate, low entropy): adjectives as part of a proper nouns. 
    Adjectives in official institutions and titles are typically capitalized and annotated as \texttt{PROPN}  in English (e.g., \textit{Secretary General}) but lowercased and annotated as \texttt{ADJ} in French (e.g., \textit{secr\'etaire g\'en\'eral}).
    }
    \label{fig:quadrant}
\end{figure}

We proceed to conduct a comparative analysis of MT vs HT based on the fine-grained morphosyntactic patterns defined in the previous section.
For any given source pattern $p$ \revb{according to the \defA{} or \defB{} definition as detailed in the previous section}, we study the distribution of its aligned target patterns, i.e.,
\htprob and \mtprob, along two major dimensions:
diversity/uncertainty as measured by entropy of the target pattern, and convergence/divergence rate. Figure~\ref{fig:quadrant} shows that there is considerable
variance in how the most frequent source patterns in HT are distributed along these two axes, and that each 
dimension captures a different property of the distribution.

Through analyses on both the aggregate level and the individual pattern level, we conclude that MT is more \emph{conservative} than HT, with less morphosyntactic diversity, more convergent patterns, and more one-to-one alignments.
We also observe that MT tends to be less similar to HT for the less frequent source patterns. 
\revb{The analyses in this section are based on the held-out subset consisting of one million sentence pairs. We refer readers to Appendix~\ref{appendix:labse} for similar results on a subset that is further filtered using LaBSE crosslingual embeddings~\citep{feng-etal-2022-language} with a remarkably similar trend, which we include to show that it does not change our conclusions when we test on data that has been filtered to improve its cross-lingual equivalence.}

\subsection{MT is Less Morphosyntactically Diverse Than HT}

\paragraph{Preliminaries}
We define diversity score as the conditional entropy of target patterns given
source patterns, which reflects the \emph{aggregate} level of uncertainty
when translating a morphosyntactic pattern.
\revb{More formally, let $P$ and $Q$ denote the categorical random variables
for source patterns and their aligned target patterns, respectively.
The aggregate diversity score is defined as
\begin{align}
    H(Q \mid P) = \sum_{p} \Pr(p)\cdot H(Q \mid P = p), \label{eq:div}
\end{align}
where $p$ is any specific source pattern that occurs in the corpus.
}

In addition, for any given source pattern $p$,
we define a \emph{source pattern-specific} diversity score
as the entropy of the target patterns aligned to that source pattern $p$. \revb{This score corresponds to the term $H(Q \mid P=p)$ in Equation~(\ref{eq:div}).}

\paragraph{Aggregate Finding}
As summarized in Table~\ref{tab:aggregate_scores} part (i), MT is less morphosyntactically diverse than
\revb{HT} in aggregate, across three language pairs and two \revb{types} of divergence.
The relative reduction in diversity for MT compared to HT ranges from 5.9\%
for \pair{Zh} (2.77 vs 2.95 with \defA{} patterns) to 22.2\%
for \pair{Fr} (1.75 vs 2.24 with \defB{} patterns).
Interestingly, {\pair{Zh}} has noticeably higher diversity scores than
\pair{Fr} and \pair{De} but lower overall reduction.
This may be attributed to the larger linguistic difference between Chinese and English.\footnote{Note that, however, our setup is
    not entirely comparable across language pairs
    since the data is not multi-way parallel.}


\begin{table}[tb]
    \centering
    \small
    \begin{tabular}{lrrrrrr}
        \toprule
         \mr{2}[2]{*}{Target} & \mc{3}{c}{\defACap{}} & \mc{3}{c}{\defBCap{}} \\
        \cmidrule(lr){2-4}\cmidrule(lr){5-7} & HT & MT & $\Delta$\% & HT & MT & $\Delta$\% \\
        \midrule\midrule \mc{7}{c}{(i) \textsc{Diversity}} \\
        \midrule
        Fr & 2.23 & 1.84 & \llap{-}17.6 & 2.24	& 1.75 & \llap{-}22.2 \\
        De & 2.23 & 1.90 & \llap{-}15.0 & 2.38 & 1.96 & \llap{-}17.8 \\
        Zh & 2.95 & 2.77 & \llap{-}5.9 & 3.79 & 3.46 & \llap{-}8.6 \\
        \midrule\midrule \mc{7}{c}{(ii) \textsc{Convergence Rate}} \\
        \midrule
        Fr & 37.9 & 44.7 & \revb{18.1} & 46.3 & 53.2 & 15.0 \\
        De & 45.8 & 51.8 & \revb{13.7} & 51.8 & 57.8 & 11.7 \\
        Zh & 21.2 & 22.6 & 6.9 & 23.4 & 25.2 & 7.4 \\
        \bottomrule
    \end{tabular}
    \caption{Aggregate diversity scores and convergence rates. The $\Delta$\% columns show the relative change in percentage from HT to MT.}
    \label{tab:aggregate_scores}
\end{table}


\begin{figure}[t]
    \centering
    \includegraphics[width=\linewidth]{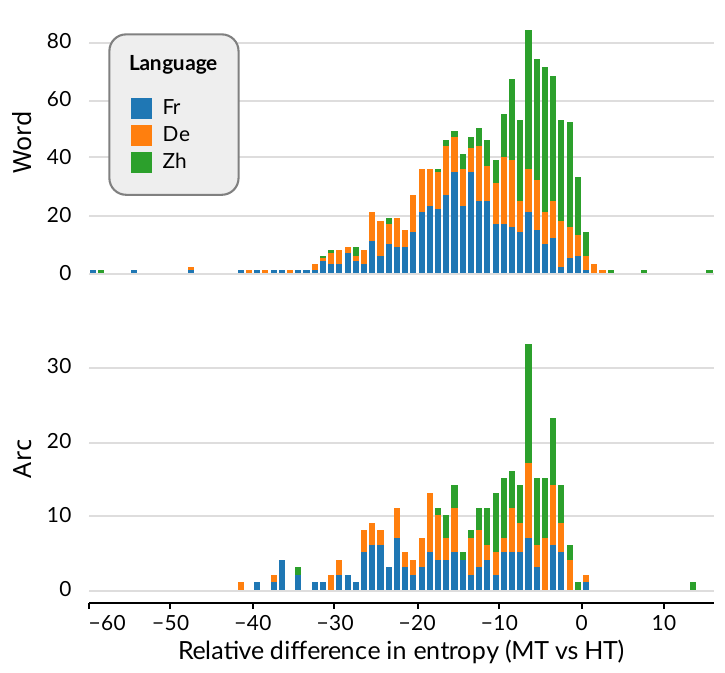}
    \caption{Stacked histogram of the relative differences in source pattern-specific diversity score.}
    \label{fig:diversity_by_pattern}
\end{figure}

\paragraph{Finding by Source Pattern}
On the level of individual source patterns, we observe that the reduction of diversity
among their aligned target patterns is \emph{across-the-board} but
\emph{unevenly distributed}.
Figure~\ref{fig:diversity_by_pattern} plots a stacked histogram of the relative differences in diversity score (MT vs HT)
for the most frequent source patterns \revb{with at least 1000 occurrences}, and it shows that the vast majority of them see a drop of diversity (i.e., negative difference). 
This reduction varies from pattern to pattern,
ranging from 0\% to 60\%.

\subsection{MT is More Convergent Than HT}


\paragraph{Preliminaries}
We tally divergences and convergences according to the two \revb{types} detailed in
Section~\ref{sec:experimental_setup}.
\revb{We then define the convergence or divergence rate as the percentage of convergent or divergent patterns out of all translation patterns.}
Similar to diversity, we can compute convergence/divergence rates for both the entire corpus in aggregate and individual source patterns.
\revb{For the latter case, we tally all the aligned target patterns for a specific source pattern and calculate the rates accordingly.}

\newcommand{\Da}{$\Delta_\text{abs}$}
\newcommand{\Dr}{$\Delta_\text{rel}$}

\paragraph{Aggregate Finding}
As summarized in Table~\ref{tab:aggregate_scores} part (ii), we observe a consistent increase
of convergence rate for all three language pairs and two \revb{types} of
divergence.
This increase is most pronounced for \pair{Fr} and \pair{De},
whereas \pair{Zh} has a less noticeable although still consistent
increase and starts with a much lower convergence rate for HT:
the highest rate for \pair{Zh} is 23.4\%, whereas \pair{De} can reach 57.8\%.

\begin{figure}[tb]
    \centering
    \includegraphics[width=\linewidth]{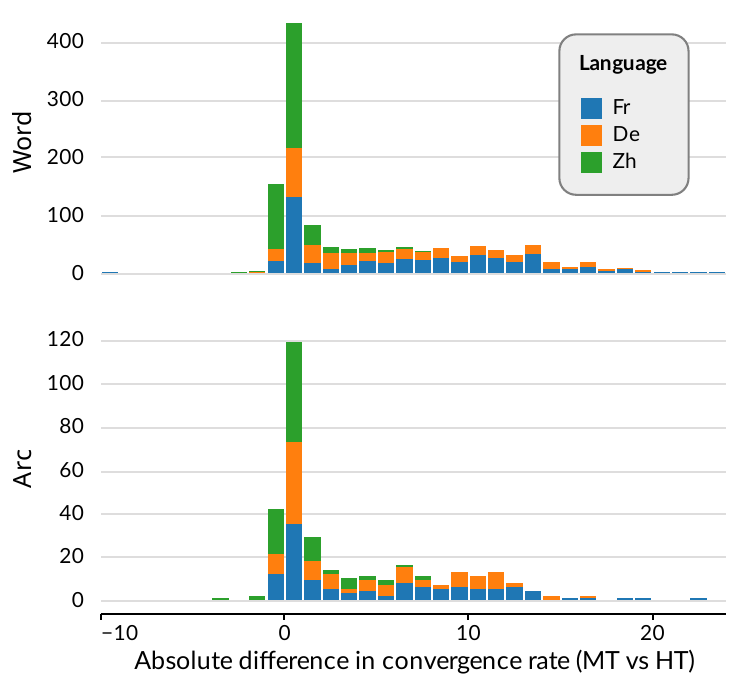}
    \caption{Stacked histogram of the absolute differences in source pattern-specific convergence rate.}
    \label{fig:div_rate_by_pattern}
\end{figure}

\paragraph{Finding by Source Pattern}
On a more granular level, we again notice a consistent increase of convergent patterns
for MT among the top source patterns (Figure~\ref{fig:div_rate_by_pattern}).
For the vast majority of top source patterns, MT has produced more convergent
translations than HT, and this discrepancy ranges from a negligible amount
(\textasciitilde 0\%) for most patterns to more than 20\%.
This discrepancy is distributed differently for the three languages:
\pair{Fr} and \pair{De} have seen more patterns with increased convergence
rate while \pair{Zh} has most patterns barely changed and clustered around 0\%.
As we later show in Figure~\ref{fig:diff_div_rate}, this trend is unsurprising
given the much lower convergence rates for \pair{Zh} in general.

\subsection{MT Looks Less Like HT For Less Frequent Patterns}
\paragraph{Preliminaries}
\rev{Both diversity score and convergence rate are properties of translations produced by one system, \emph{either} MT \emph{or} HT.
To directly measure the distributional difference between MT and HT, we resort to} Wasserstein distance (WD) between the two conditional distributions\footnote{\revb{Recall that we treat both source patterns and target patterns as categorical random variables where every unique source or target pattern is treated as a distinct value that the random variables can take.}} \mtprob and \htprob using a unit cost matrix.\footnote{
In which diagonal/off-diagonal entries are 0/1.}
This metric can be intuitively interpreted as the minimal amount of probability mass that has to be moved from \mtprob to match \htprob, with an upper
bound of 1 (i.e., sum of all probability mass).\footnote{We note that other metrics such as KL-divergence can also be used to measure distributional difference, but we eventually chose WD for its interpretability.}

\begin{figure}[t]
    \centering
    \includegraphics[width=\linewidth]{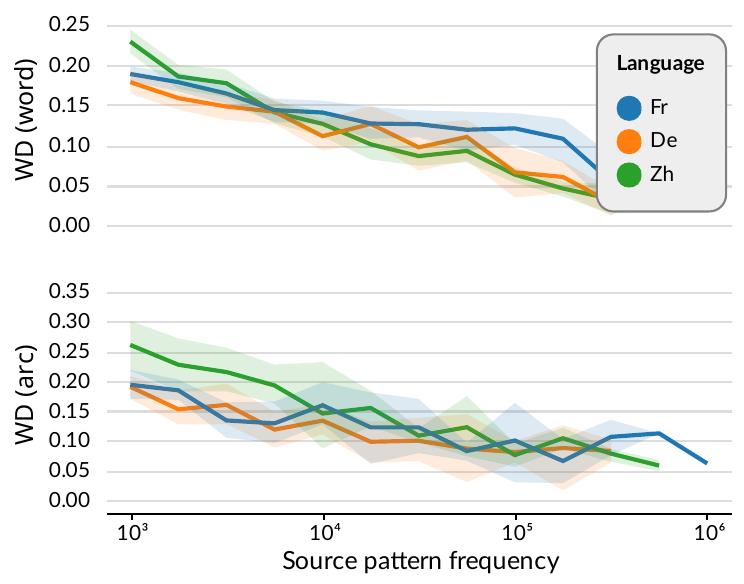}
    \caption{Wasserstein distance with a unit cost matrix between \mtprob and \htprob for any given source pattern $p$.
    Patterns are binned by frequency on a log scale, and both the means (lines) and the 95\% confidence intervals (shaded areas) are shown.
    The plot shows a negative correlation between WD and the source pattern frequency.
    }
    \label{fig:wasserstein_distance}
\end{figure}

\paragraph{Finding}
As Figure~\ref{fig:wasserstein_distance} shows,
there is a negative correlation between WD and the source pattern frequency: 
MT matches HT more closely for the more frequent source patterns while having difficulty
in reproducing the HT distribution for the less frequent ones.
This trend persists for all tested settings, and points to a potential weakness of MT systems when it comes to learning the distributions of the less common structures.

\subsection{Beyond One-to-one Alignments}

\paragraph{Preliminaries}
One-to-one alignments constitute a majority of all detected alignments, but they
fail to account for translation patterns involving deletions and insertions.
To investigate the quantitative differences between HT and MT on those special
patterns, we conduct additional analyses on the distribution of all categories
of alignments based on the \defA{} definition.
Besides deletions (\texttt{src2null}) and
insertions (\texttt{null2tgt}), the remaining alignments
are collapsed into the \texttt{other} category (e.g., one-to-many mapping).

\begin{figure}[tb]
    \centering
    \includegraphics[width=\linewidth]{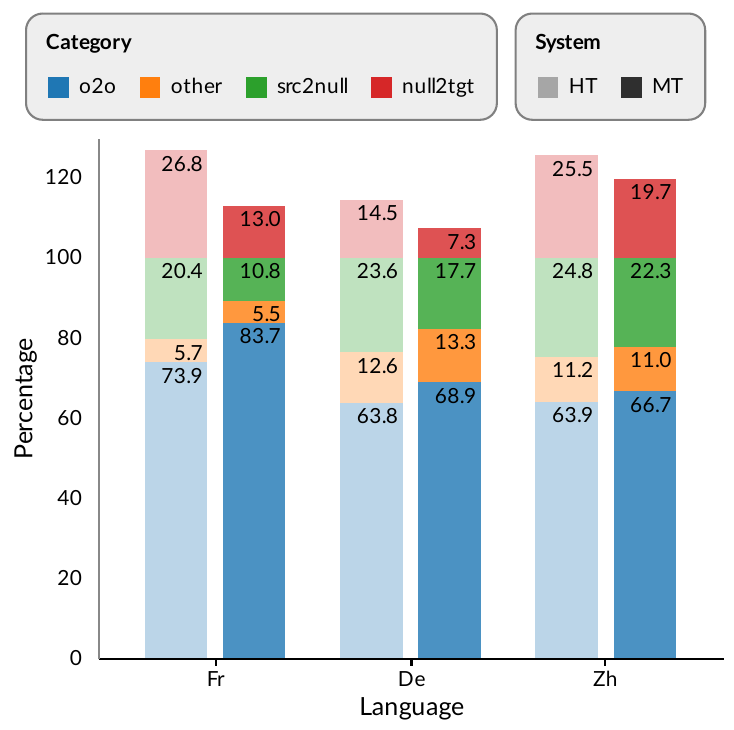}
    \caption{Distribution for all types of alignments. Percentages are defined
        relative to the total number of source content words.
        \texttt{o2o}: one-to-one;
        \texttt{src2null}: deletions; \texttt{null2tgt}: insertions; \texttt{other}: other types such as
        one-to-many.}
    \label{fig:null}
\end{figure}

\paragraph{Finding}
Figure~\ref{fig:null} summarizes the distribution of all alignment categories,\footnote{The percentages are computed
    in terms of source words. By definition, \texttt{src2null}, \texttt{o2o}
    and \texttt{other} add up to 100\%. Since \texttt{null2tgt} alignments do not have
    aligned source words, their percentages indicate how many target content words are inserted
    for each content word on the source side.}
which demonstrates a significant and consistent difference between HT and MT.
More specifically, MT produces fewer deletions (green), fewer insertions (red),
and more one-to-one translations (blue).
\pair{Fr} again exhibits the biggest discrepancy with 9.6\% less deletions (10.8\% vs 20.4\%)
and 14.8\% less insertions (13.0\% vs 26.8\%), both around 50\% relative reduction.
This trend contributes to the overall conservative nature of MT predictions, favoring one-to-one alignments at the expense of the other (more uncertain) categories. 


\section{Understanding the Discrepancy}
\label{sec:understanding}

In this section, we seek to understand the source of
discrepancy between HT and MT as demonstrated in the previous section.
By investigating different decoding algorithms, we attribute this discrepancy
to the use of beam search, echoing the thesis laid out by
previous work~\cite{edunov-etal-2018-understanding,eikema-aziz-2020-map}.
\revb{More specifically in our experiments, we show that beam search is biased towards less diverse and more
convergent translations, even when the learned model distribution actually resembles HT. This bias is most prominent when the convergent
patterns \biasDescPlural{}. 
Moreover, frequencies of convergent patterns in MT are increased even when they are uncommon in HT, suggesting perhaps a more inherent structural bias in current MT architectures.}

\paragraph{Decoding Algorithms}
Besides beam search, we additionally obtain translations through
two sampling methods.
More specifically, to make fair comparison with single-reference HT,
we sample one translation using ancestral sampling or nucleus
sampling with $p = 0.95$~\cite{Holtzman2020The} for each source sentence.

\begin{figure}[tb]
    \centering
    \includegraphics[width=\linewidth]{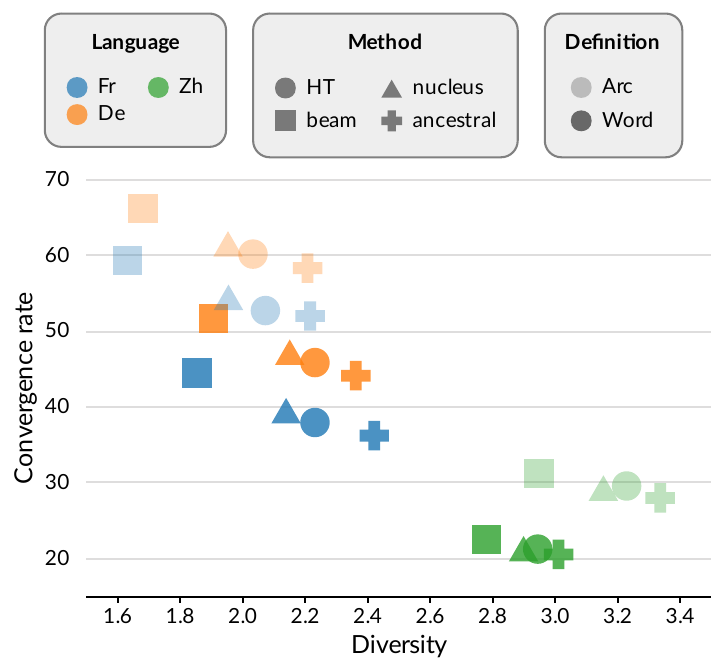}
    \caption{Convergence rates (Y axis) and diversity scores (X axis) on the aggregate level for translations through different sampling methods
        and HT. Sampling methods consistently obtain higher diversity score and lower
        convergence rate than beam search.}
    \label{fig:sampling}
\end{figure}

\paragraph{Beam Search is Biased Against Diversity and Divergence}
As Figure~\ref{fig:sampling} illustrates, for all three language pairs and
two \revb{types} of divergence,
translations obtained through beam search are
significantly less diverse and more convergent compared to either sampling method.
Indeed, ancestral sampling consistently produces higher diversity scores and lower
convergence rates than even HT.\footnote{We hypothesize that the increased diversity score
    and the higher divergence rate for
    ancestral sampling compared to HT are attributable to the use of
    label smoothing during training. \citet{Roberts2020} have also demonstrated the effect of label smoothing on various diversity diagnostics.}
Since ancestral sampling is an unbiased estimator of the model distribution,
this suggests that on the aggregate distribution level,
the model learns to be as least as morphosyntactically diverse and divergent as HT.

\begin{figure}[tb]
    \centering
    \includegraphics[width=\linewidth]{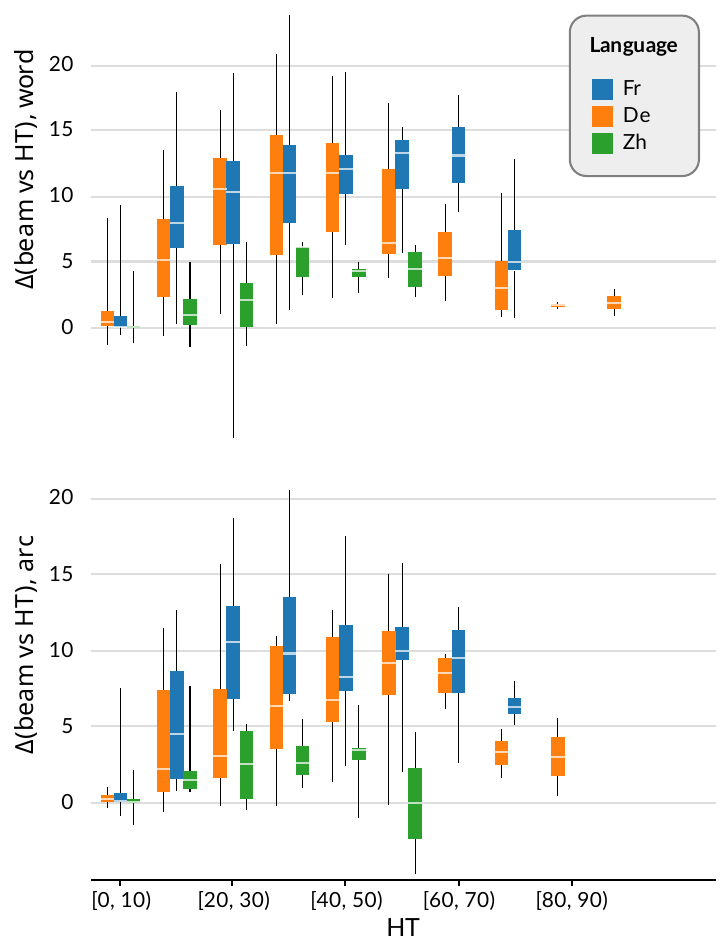}
    \caption{Plot of difference in convergence rate (beam search vs HT)
        against convergence rate of HT. The plot is similar when comparing beam search against ancestral sampling.}
    \label{fig:diff_div_rate}
\end{figure}

A further breakdown of \rev{most frequent\footnote{With at least 1000 occurrences.}} individual source patterns reveals that beam search's bias
towards convergent translations is \rev{a function of} the relative frequencies of the
convergent patterns.
As Figure~\ref{fig:diff_div_rate} demonstrates,
the increase of convergence rate for beam search compared to ancestral sampling
seems to be quadratically correlated with the convergence rate for ancestral sampling:
Peak difference is reached at around 40-50\%.
This suggests that beam search favors the convergent pattern more when the pattern \biasDescSingular{}. \rev{This could be because the model has seen the pattern enough to assign it substantial probability mass, but there is still enough uncertainty that humans will frequently choose other patterns.}

We additionally note that convergence rate increases
for the overwhelming majority of the most frequent source patterns 
\emph{even when the 
convergence patterns are uncommon in HT}.
This strongly suggests an inherent bias of beam search towards convergent patterns,\footnote{We do not observe a similar trend when comparing ancestral sampling against HT.}
and that this bias is distinct from the typical 
bias amplification due to data exposure, e.g., ``cooking'' is more likely to co-occur with ``women'' than ``men'' in the training data~\cite{zhao-etal-2017-men}.
We suspect that this bias towards convergence is due to the architectural design of MT systems, but we leave the subject matter for future work.

\section{Divergence and MT Quality}
\label{sec:difficulty}

In our final analysis, we investigate how the presence of morphosyntactic divergence
in HT might affect MT quality.
In contrast to the previous sections analyzing conditional distributions given a source pattern, we focus instead on individual divergences/convergences.
The potential connection between divergence and MT quality is motivated by 
second-language acquisition research
that describes language inference
from their first languages (i.e., negative transfer)
as one source of difficulty for learners~\cite{gass2020second},
which can happen when the two languages diverge structurally.
Do MT systems have similar problems with divergences?

\newcommand{\ctrlG}{control group}
\newcommand{\expG}{experiment group}

\paragraph{Preliminaries}
To answer this question, we conduct an analysis on the presence (or absence) of
a \defA{} morphosyntactic divergence in HT and the corresponding 
MT quality as measured by BLEU~\cite{papineni-etal-2002-bleu}
and BLEURT~\cite{sellam-etal-2020-bleurt}.
\revb{The basic idea is to construct two contrastive groups of source sentences (called the \expG{} and the \ctrlG{}) and compare the MT performance on each group. The HT references of the \expG{} contain a given divergent pattern, corresponding to sentences that are perhaps more challenging to translate, whereas those of the \ctrlG{} do not.}

More specifically, for a given divergence with source pattern $p$ and target pattern $q$ $(p \neq q)$, its \ctrlG{} consists of source sentences for which HT translates every source $p$ into target $p$ (i.e., a convergent pattern), and its \expG{} \revb{consists of source sentences for which HT translates every source $p$ into $p$ except for one that is translated into $q$. For an simplified example, if we are interested the divergence that translates nouns into verbs, the corresponding \ctrlG{} contains source sentences for which HT translates every noun into a noun, whereas its \expG{} contains source sentences for which exactly one noun is translated into a verb and the rest of nouns into nouns.}

We then \revb{collect the MT outputs} for both groups and compute the differences in BLEU 
and BLEURT. \revb{This procedure is repeated for every divergence pattern}
for which both groups have at least 100 sentences.

\begin{figure}
    \includegraphics[width=\linewidth]{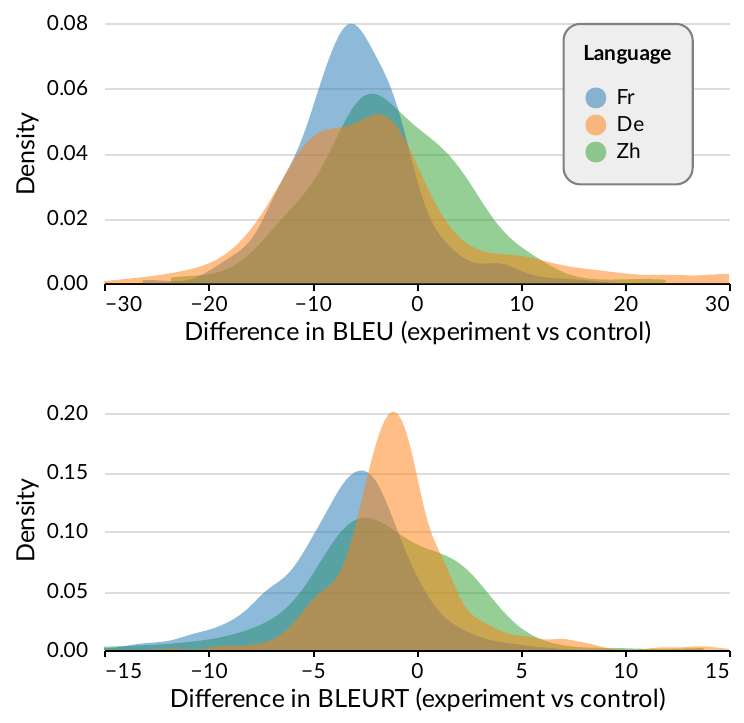}
    \caption{Kernel density estimation for the difference in BLEU or BLEURT scores
        between the \expG{} and the \ctrlG{}. Negative values indicate that the
        \expG{} has lower score than the \ctrlG{}.}
    \label{fig:corr_density}
\end{figure}

\paragraph{Findings}
\revb{We treat each difference in BLEU or BLEURT as one data point and plot their estimated probability density function.} As illustrated in Figure~\ref{fig:corr_density}, divergences are more often
associated with significantly lower BLEU scores (i.e.,
negative differences),
with a fairly large amount of variance.
Trends for BLEURT scores are similar, but with \pair{De} showing less drastic differences compared to BLEU.\footnote{We also note that ngram overlap-based metrics such as BLEU are more likely to penalize diverse translations~\cite{freitag-etal-2019-ape}.}
On the other hand, a substantial number of divergent patterns have either
virtual no change or an increase of BLEU or BLEURT scores.
This suggests that being a divergence pattern in itself is not associated
with decreased MT performance.

What could explain this variance?
Why are some divergent patterns associated with worse MT performance \rev{while others aren't}?
One obvious hypothesis is that these patterns are seen less frequently
during training. However, a closer inspection seems to suggest that frequency
of \rev{divergent} patterns alone is not an adequate predictor.
More specifically, we \rev{use the absolute or relative frequency\footnote{\rev{Here, relative frequency is defined as the ratio of the number of training examples with the divergence over that with the convergence. It is a way to counterbalance the fact that some extremely common source patterns will have a lot more frequent divergences.}} of the divergent pattern, with or without taking a log of the number, and correlate it with BLEU or BLEURT scores. Even with the best option (log of relative frequency) presented in Table~\ref{tab:ratio_corr}, there is only weak correlation (Pearson or Kendall $\tau$) for \pair{Fr} and \pair{De}, and no correlation for \pair{Zh}}.
It is unclear what aspects of divergent patterns make them
more difficult to translate, or whether they are merely co-occurring with
those elements that are the true cause of difficulty.
We leave it to future work to investigate the underlying cause.

\newcommand{\pv}[1]{{\tiny\color{darkgray} #1}}
\begin{table}[tb]
    \centering
    \footnotesize
    \begin{tabular}{llr@{\hskip 0.15cm}lr@{\hskip 0.15cm}l}
        \toprule
        Target              & Metric & \mc{2}{c}{Pearson} & \mc{2}{c}{Kendall $\tau$}                         \\
        \midrule
        \multirow{2}{*}{Zh} & BLEURT & -0.072             & \pv{0.32}                 & 0.030  & \pv{0.54}    \\
                            & BLEU   & -0.080             & \pv{0.27}                 & -0.026 & \pv{0.59}    \\
        \midrule
        \multirow{2}{*}{Fr} & BLEURT & 0.319              & \pv{1.4e-16}              & 0.240  & \pv{1.0e-19} \\
                            & BLEU   & 0.206              & \pv{1.6e-7}               & 0.161  & \pv{1.2e-9}  \\
        \midrule
        \multirow{2}{*}{De} & BLEURT & 0.289              & \pv{1.4e-11}              & 0.253  & \pv{3.6e-18} \\
                            & BLEU   & 0.159              & \pv{2.5e-4}               & 0.171  & \pv{4.4e-9}  \\
        \bottomrule
    \end{tabular}
    \caption{Correlation between the difference in BLEURT score and ratio of frequencies (i.e.,
        the number of training examples with divergences over that with convergences).
        $p$-values are displayed in gray.}
    \label{tab:ratio_corr}
\end{table}

\section{Conclusion}
\label{sec:conclusion}
We conduct a large-scale fine-grained comparative investigation between HT and MT outputs, through the 
lens of morphosyntactic divergence.
Based on extensive analyses on three language pairs, we demonstrate that MT is less morphosyntactic diverse 
and more convergent than HT.
We further attribute to this difference to the use of beam search that biases MT outputs towards
less diverse and less divergent patterns.
Finally, we show that the presence of divergent patterns in HT has overall an adverse effect on MT quality.

In future work, we are interested in applying the same analysis to large language model (LLM)-based MT systems. 
Recent studies have noted that LLM-based systems tend to produce less literal translations, compared to the traditional
encoder-decoder models~\cite{vilar-etal-2023-prompting,raunak-etal-2023-gpts}.
It would be interested to see whether and to what extent the LLM translations might differ from those produced by traditional models
when viewed from a morphological lens.

\section*{Acknowledgements}
We thank Julia Kreutzer, Eleftheria Briakou, Markus Freitag and Macduff Hughes for providing useful feedback that helped shape this paper. We would also like to thank the anonymous reviewers and the action editor for their constructive comments.

\bibliography{tacl2021}
\bibliographystyle{acl_natbib}

\onecolumn
\appendix
\section{Analysis Subset Filtered Using LaBSE Embeddings}
\label{appendix:labse}

\revb{The main results of the paper are obtained on a held-out subset of the WMT data. To remove some of the noise due to the automatic extraction pipeline that produced the WMT data, we resort to LaBSE embeddings~\citep{feng-etal-2022-language} to further filter the original held-out subset. 
More specifically, we use the LaBSE model to derive the crosslingual embeddings for the source and the target of
any sentence pair, and sort all pairs based on the cosine distance between the source and the target embeddings.
The top half (i.e., lowest distance) is kept for analysis, resulting in 500K sentence pairs for 
each language pair.
}

\begin{table}[h]
    \centering
    \small
    \revb{
    \begin{tabular}{lrrrrrr}
        \toprule
         \mr{2}[2]{*}{Target} & \mc{3}{c}{\defACap{}} & \mc{3}{c}{\defBCap{}} \\
        \cmidrule(lr){2-4}\cmidrule(lr){5-7} & HT & MT & $\Delta$\% & HT & MT & $\Delta$\% \\
        \midrule\midrule \mc{7}{c}{(i) \textsc{Diversity}} \\
        \midrule
        Fr & 2.22	 & 1.85	 & \llap{-}16.8 & 2.25		& 1.77	 & \llap{-}21.6 \\
        De & 2.24	 & 1.95 & \llap{-}12.9 & 2.39	 & 2.01	 & \llap{-}16.0 \\
        Zh & 2.92	 & 2.76	 & \llap{-}5.5 & 3.78 & 3.47	 & \llap{-}8.3 \\
        \midrule\midrule \mc{7}{c}{(ii) \textsc{Convergence Rate}} \\
        \midrule
        Fr & 37.4 & 43.8 & 17.1 & 45.4 & 52.0 & 14.5 \\
        De & 44.4 & 49.4 & 11.3 & 49.6 & 54.7 & 10.3 \\
        Zh & 20.9 & 22.0 & 5.4 & 23.0 & 24.5 & 6.6 \\
        \bottomrule
    \end{tabular}
    \caption{Aggregate diversity scores and convergence rates for the LaBSE-filtered subset. The $\Delta$\% columns show the relative change in percentage from HT to MT.}
    \label{tab:aggregate_scores_labse}
    }
\end{table}

\revb{Table~\ref{tab:aggregate_scores_labse} summarizes the aggregate diversity scores and convergence rates. 
The relative changes are slightly smaller than those in Table~\ref{tab:aggregate_scores}, but the overall trend is remarkably similar: For both \defA{} and \defB{} divergences, MT produces less diverse outputs with more convergent patterns.
}

\section{Percentage of Content Words and Their Alignments}
\label{appendix:content_stats}

\begin{table}[h]
    \centering
    \small
    \revb{
    \begin{tabular}{lrrr}
        \toprule
        Lang & Source content words & Target content words & Alignments \\
        \midrule
        Fr & 13.7M / 28.7M = 47.9\% & 14.8M / 33.7M = 44.0\% & 11.3M / 27.2M = 41.7\% \\
        De & 10.2M / 21.3M = 48.1\% & 8.8M / 20.1M = 43.8\% & 7.9M / 18.5M = 42.9\% \\
        Zh & 12.6M / 26.2M = 48.2\% & 12.8M / 24.5M = 52.4\% & 10.1M / 22.3M = 45.3\% \\
        \bottomrule
    \end{tabular}
    \caption{Percentage of content words and their alignments for the held-out analysis subset.}
    \label{tab:content_stats}
    }
\end{table}

\revb{Table~\ref{tab:content_stats} summarizes the percentage of content words and their alignments based on the held-out analysis subset. We only keep the alignments for the main results if both the source token and the target token are content words. The statistics show that around 40\%-50\% of the tokens (either on the source or the target side) are considered content words, and a similar percentage of alignments pass our criterion.
}

\end{document}